\documentclass[10pt,journal,compsoc]{IEEEtran}

\ifCLASSOPTIONcompsoc
  \usepackage[nocompress]{cite}
\else
  \usepackage{cite}
\fi

\ifCLASSINFOpdf
   \usepackage[pdftex]{graphicx}
\else
\fi

\usepackage{capt-of}  
\usepackage{cuted}
\usepackage{algorithm}
\usepackage{algpseudocode}
\usepackage{CJKutf8}
\usepackage{tabularx}
\usepackage{array}
\usepackage{multirow}
\usepackage{xfakebold}
\usepackage{placeins}
\usepackage{mathtools}
\usepackage{caption}
\usepackage{subcaption}
\PassOptionsToPackage{hyphens}{url}\usepackage[hyperfootnotes=false]{hyperref}
\newcolumntype{P}[1]{>{\raggedright\arraybackslash}p{#1}}
\newcolumntype{M}[1]{>{\raggedright\arraybackslash}m{#1}}
\newcolumntype{x}[1]{>{\centering\arraybackslash\hspace{0pt}}p{#1}}

\usepackage{color}
\definecolor{myred}{rgb}{.8,.0,.0}

\begin{document}
\begin{CJK}{UTF8}{ipxm}

\title{Detection of Furigana text in images}
\author{Nikolaj Kjøller Bjerregaard, Veronika Cheplygina, Stefan Heinrich \\
IT University of Copenhagen, Copenhagen, Denmark\\
\texttt{nikolajnkb@gmail.com*, vech@itu.dk, stehe@itu.dk}
}

\maketitle

\begin{abstract}
\textit{Furigana} are pronunciation notes used in Japanese writing. Being able to detect these can help improve optical character recognition (OCR) performance or make more accurate digital copies of Japanese written media by correctly displaying \textit{furigana}.
This project focuses on detecting \textit{furigana} in Japanese books and comics. While there has been research into the detection of Japanese text in general, there are currently no proposed methods for detecting \textit{furigana}.

We construct a new dataset containing Japanese written media and annotations of \textit{furigana}. We propose an evaluation metric for such data which is similar to the evaluation protocols used in object detection except that it allows groups of objects to be labeled by one annotation.
We propose a method for detection of \textit{furigana} that is based on mathematical morphology and connected component analysis.
We evaluate the detections of the dataset and compare different methods for text extraction. We also evaluate different types of images such as books and comics individually and discuss the challenges of each type of image.

The proposed method reaches an F1-score of 76\% on the dataset. The method performs well on regular books, but less so on comics, and books of irregular format. Finally, we show that the proposed method can improve the performance of OCR by 5\% on the manga109 dataset.

Source code is available via 
\texttt{\url{https://github.com/nikolajkb/FuriganaDetection}}
\footnote{This project was originally submitted by NKB in fulfillment of the 30 ECTS MSc thesis at the IT University of Copenhagen.}

\end{abstract}

\section{Introduction}\label{sec:introduction}

\begin{figure}
\centering
\includegraphics[width=7cm]{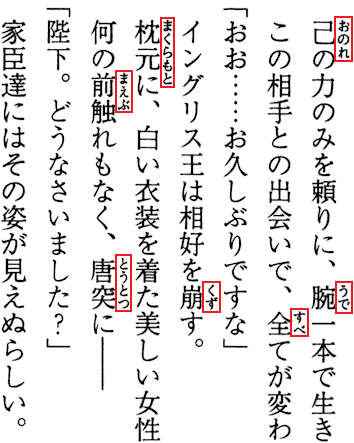}
\caption{\textit{Furigana} marked by red boxes on an excerpt from a book. Source: ``Eiyū-Ō, Bu o Kiwameru Tame Tensei-Su: Soshite, Sekai Saikyō no Minarai Kishi'' by Hayaken \cite{eiyou}.}
\label{furigana_book}
\end{figure}

\begin{figure}
\centering
\includegraphics[width=8.5cm]{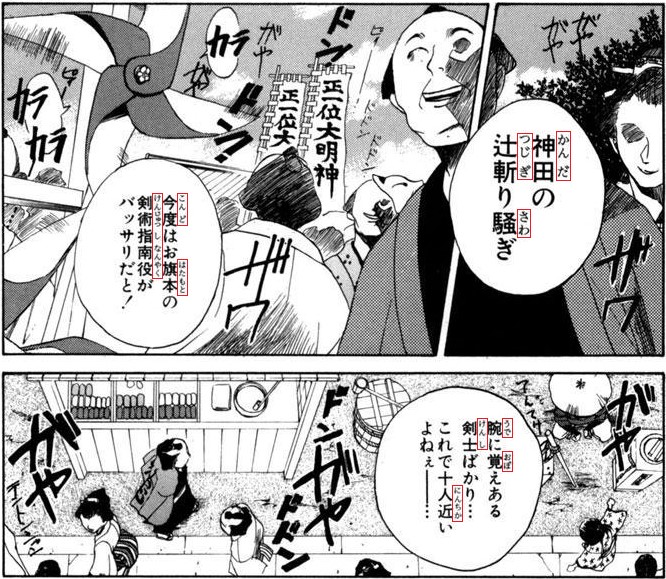}
\caption{\textit{Furigana} marked by red boxes on a comic book page. 
Using Tesseract OCR on the top left speech bubble gives the following text:\\
こんとと 公才はお誠本の 本和導が バッサリだと/.\\ 
Removing the \textit{furigana} from the image and running OCR again gives the following text:\\ 
今度はお旗本の 剣術指南役が パッサリだと/.\\ 
The version without \textit{furigana} is a near-perfect transcription of the image (edit distance 2), while the version with \textit{furigana} shares little similarity at all with the original text (edit distance 20). Source: ``Akera Kanjinch'' by Yuki Kobayashi\cite{akkera}.}
\label{furigana_comic}
\end{figure}

\IEEEPARstart{F}{urigana} is a part of Japanese written language. Japanese uses both a phonetic (representing sounds, called \textit{hiragana}) alphabet and a logographic (representing meaning, called \textit{kanji}) alphabet. In written Japanese, the two are mixed to form sentences. For \textit{kanji}, since the characters represent meaning, the reader may not always know how they are pronounced. Names is an example where the reading of \textit{kanji} is particularly difficult, given that names written with the same characters can often be pronounced many different ways as described by Ogihara in \cite{ogihara2021know}. Therefore, writers may sometimes add notes next to \textit{kanji} to indicate their pronunciation, these types of notes are called \textit{furigana}. \textit{Furigana} is typically written in the \textit{hiragana} alphabet.
Fig. \ref{furigana_book} shows \textit{furigana} on a book page, Fig. \ref{furigana_comic} shows furigana on a comic book page. 

\textit{Furigana} can be problematic for systems that process text within images. \textit{Furigana} does not change the meaning of the text and can thus be disregarded by computers for most purposes. For example, current OCR systems do not handle \textit{furigana} well. The \textit{furigana} is often mistaken as regular text and inserted into the output or characters are misclassified because of the \textit{furigana} next to them. An example of this is described in Fig. \ref{furigana_comic}.

Detecting the locations of \textit{furigana} could also be used to make more accurate digital copies of scanned texts. Most data formats for displaying text such as pdf, epub, and html has support for displaying \textit{furigana}. By detecting what text is \textit{furigana}, \textit{furigana} could be displayed correctly in these formats.

The proposed system aims to detect the location of \textit{furigana}, so that other systems may use this information to better process Japanese text within images. \textit{Furigana} rarely appears ``in the wild'', such as on signs or advertisements. This is not to say that such cases do not exist, but they are rare, and there is currently no dataset containing such images. Additionally, \textit{furigana} is generally not used in handwritten text. This project therefore focuses on two categories: books and comics. 


\section{Related work}\label{sec:relatedwork}

There are currently no works directly treating the detection of \textit{furigana}. However, several papers that investigate related problems: text analysis in images (OCR, scene text detection, and the use of mathematical morphology for detecting text in images), and then methods specifically designed for Japanese text (handwritten text and text in Japanese comics). 

\subsection{Text analysis in images}

\subsubsection{Optical character recognition}
The problem of locating \textit{furigana} is related to the more general problem of text localization. Text localization is a fundamental step in most systems that processes text in images. This includes OCR systems such as Tesseract \cite{smith2007overview}, which uses connected component analysis to locate lines of text before splitting said lines into individual characters. In newer versions of Tesseract, a long short-term memory (LSTM) network based approach has also been introduced. OCRopus\cite{breuel2008ocropus} starts by analyzing the physical layout of the page such as columns, text blobs, and lines to find the location of text. PaddlePaddleOCR(PP-OCR)\cite{du2020pp} is another OCR system that is of particular interest for this project since it has placed a lot of focus on the processing of Chinese text, which is more comparable to Japanese than English. PP-OCR uses differentiable binarization as described by Liao et. al. \cite{liao2020real} for text localization. This method involves using a probabilistic segmentation map of the text in the image to estimate the bounding boxes for the text.

\subsubsection{Scene text}
Recently a more difficult challenge known as scene text detection has been widely studied. Historically, OCR systems mainly dealt with recognizing text in machine-written documents. Scene text deals with text that is captured "in the wild", that is, not in written media, but rather on images of street signs, billboards, storefronts, and more. Scene text is more difficult to process due to complex backgrounds, noise, warped perspective, and other challenges that do not appear in machine-written documents. Solutions to the scene text detection need to handle a problem that also shows up when processing text in comic books: how to find text in an image with many visual distractions? 

Many datasets containing scene text have been created, such as COCO-Text \cite{veit2016coco} which is based on the MSCOCO \cite{lin2014microsoft} object detection dataset created by Microsoft, to which researchers from Cornell Tech have extracted images containing text and added annotations for the text. ICDAR \cite{karatzas2015icdar, karatzas2013icdar, shahab2011icdar} is created for the International Conference on Document Analysis and Recognition and has multiple versions containing scene text of different types. Total Text \cite{ch2017total} by Ch’ng et. al. puts special focus on curved text. French Street Name Signs (FSNS) \cite{smith2016end} contains more than one million images of French road signs from Google Street View. There is even a dataset containing Japanese text, downtown Osaka scene text \cite{iwamura2016downtown} which was captured on a 360 degree camera and thus has many angles for each text instance. 

The Robust Reading Competition is held every year and focuses on ``written communication in unconstrained setting'', which often includes scene text detection \cite{karatzas2018robust}. Current top text detection methods include: 
\begin{itemize}
    \item TextFuseNet \cite{ye2020textfusenet}, which uses Concurrent Neural Network (CNN) based networks in a multi-path fusion architecture to exploit character, word, and global level features
    
    \item Corner-based Region Proposal Network \cite{deng2019detecting}, which uses detection of word-corners to predict bounding boxes
    \item EAST \cite{zhou2017east}, a popular text detection system that uses a fully convolutional network to detect text locations.
    \end{itemize}

\subsubsection{Morphology for text detection}
Several papers use morphology and connected component analysis for locating text. Qi et al. \cite{qi2005using} present 12 features for text that can be found using connected component analysis and use Adaboost \cite{freund1997decision} to find scene text using these features. Wu et al. \cite{wu2008morphology} use a similar approach of feature extraction and classification to detect text in scene text images. Dos Santos et al. \cite{dos2009text} use morphology to extract features from pages containing handwritten text which are used to predict the locations of text lines.

\subsection{Japanese text}

\subsubsection{Handwritten Japanese}
Many of the papers that treat Japanese text detection only consider handwritten text. Zhu et. al.\cite{zhu2010robust} describe a probabilistic model for recognizing text from the HANDS-kondate dataset of handwritten Japanese text \cite{MatsushitaKondate}. Ly et. al. \cite{ly2018training} present an end-to-end deep convolutional recurrent network trained on a dataset of synthetic handwritten text. Clanuwat et. el. \cite{clanuwat2018deep} focus on recognizing ancient Japanese text using deep learning. Finally, Ly et. al. \cite{ly2020attention} focus on historical handwritten text, creating an attention-based row-column encoder-decoder model for recognizing such text. Despite dealing with handwritten text as opposed to machine-written text, these papers do highlight some of the general issues related to processing Japanese text in images, such as text being written in multiple orientations and the difficulty of segmenting Japanese characters.

\subsubsection{Text detection in Japanese comics}
There has also been research into processing text specifically in Japanese comics (also known as manga). A key dataset for much of the ongoing research is Manga109 \cite{fujimoto2016manga109,aizawa2020building}. This dataset contains 109 different manga and around 20,000 pages in total. The dataset provides annotations which includes the locations of frames, characters, and text along with transcriptions of the text. Mantra Inc. has created a tool for automatic translation of manga, their paper discusses many of the key aspects in the processing of manga \cite{hinami2020towards}. Although they mainly discuss methods for translation, the problem of localizing text is also mentioned. Their method involves an object detector for speech bubbles and one for text lines. Methods specifically for text localization in manga have also been proposed, such as \cite{aramaki2016text,chu2018text,tolle2011method,piriyothinkul2019detecting} and especially good results from \cite{del2020unconstrained}. We discuss these in more detail in Section \ref{sec:method:extracting}.


\section{Problem analysis}

This section analyses the problem of detecting \textit{furigana} by detailing how \textit{furigana} is used in Japanese and how it appears in written media. This information gives a general idea of what defines \textit{furigana} as well as some pitfalls one may encounter when trying to locate it. 

\subsection{Characteristics of furigana}\label{sec:furigana}
To understand how to solve the problem of locating \textit{furigana}, we can begin by analyzing some characteristics of how \textit{furigana} appears within Japanese text. The text shown in Fig. \ref{furigana_book} and \ref{furigana_comic} is read from right to left and from top to bottom, that is to say, the lines of text are written vertically. This is the case for the majority of written media in Japan. However, Japanese text can be written both vertically and horizontally. For example, most digital text, such as on websites and in video games is written horizontally. In addition, short texts are often written horizontally, such as text on billboards or restaurant menus.
\par
When the text is written vertically, the \textit{furigana} will appear on the right side of the \textit{kanji} that it is annotating. If the text is written horizontally, the \textit{furigana} is usually written on top of the \textit{kanji} or in a few cases on the bottom. If a word is made up of more than one kanji, the \textit{furigana} may appear in one continuous string or with a space separating the \textit{furigana} based on which \textit{kanji} it is annotating. \textit{Furigana} is always written in a smaller font than the main text. 
\par
\textit{Furigana} is almost always written in the \textit{hiragana} alphabet, but this is not always the case. Sometimes, the author may write \textit{furigana} in the \textit{katakana} alphabet, which is a second phonetic alphabet used in Japanese, mainly for foreign words. In some cases, an author might even write "\textit{furigana}" using \textit{kanji}. For example, an author might write ガール, a transliteration of the English word 'girl' and annotate it with 女子, the Japanese word for girl. This type of usage does not fit most definitions of \textit{furigana}, but in this project we shall consider it part of the text that should be detected since it is still just a type of reading aid.

\subsection{Challenges of Japanese written media}\label{sec:challenges}

\begin{figure}
\centering
\fbox{\includegraphics[width=85mm]{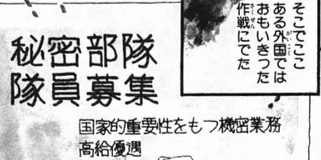}}\hfill
\\\vspace{0.5cm}
\fbox{\includegraphics[width=40mm]{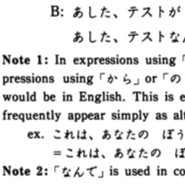}}\hfill
\fbox{\includegraphics[width=41mm]{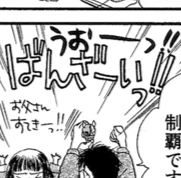}}
\caption{Challenging texts. Top: here, different font sizes and orientations are mixed on a single page. Src:\cite{peppermint}. Bottom left: English and Japanese text mixed together in a textbook. Src:\cite{jpfor} Bottom right: text with special formatting. Src:\cite{derby}}
\label{fig:challenges}
\end{figure}

Any solution to the problem of detecting the location of \textit{furigana} must consider challenges imposed by the nature of the data in question. Failing to take these challenges into account will mean that the system does not generalize to a diverse set of data. We provide a list of challenges, which is result of discoveries made throughout this project as well as NKB's experience reading Japanese books and comics, below. Some examples are shown in Fig. \ref{fig:challenges}.

\textbf{Images may contain more than just text}. For regular books, pages almost exclusively consist of text, but that is not to say that other elements do not occur. Books may include images, graphs, or shapes used for formatting such as boxes, lines, circles, or arrows. As for comics, these contain a large number of visual elements, making it vital that a solution can deal with this complexity.

\textbf{Images may be of different resolutions}. The physical size of books differs, but even for books of the same size, the resolution of the scanned pages varies greatly. This means that relying on pixel-based sizes may not generalize well.

\textbf{Differences in font sizes}. Different books use different font sizes. Even a single page may contain multiple font sizes. One cannot assume that a single page only has one font size. In books, headers are often larger than other text, and in comics, the font size might vary from one speech bubble to another.

\textbf{\textit{Furigana} is small}. \textit{Furigana} is usually written in very small font sizes. Depending on the quality of the scan, it might be difficult even for a human to tell what characters are written.

\textbf{Images may be rotated}. Sometimes an image will not have the correct orientation after being scanned.

\textbf{Images may contain slanted, warped and curved text}. Although most text in written media is written in straight lines, slanted, warped or curved text might appear. Particularly, in Japanese comics it is common for the author to occasionally include handwritten text, this text is often not written in a straight line. It should be noted that it is uncommon for such text to include \textit{furigana}.

\textbf{Text might not be black}. Although rare, text might be written in a font that is not black, such as white text on a black background. 

\textbf{Different text orientations on a single page}. Japanese can be written horizontally and vertically. Additionally, there might be both horizontal and vertical text on the same page.

\textbf{Pages may contain text that is not Japanese}. Particularly, small bits of English text can sometimes be found in Japanese books. In language textbooks for Japanese learners, much of the text is often not Japanese.

\section{Dataset}
The dataset was created to represent a diverse selection of books and comics. It includes all the cases described in Section \ref{sec:challenges}, except for rotated pages. In this project, the dataset is only used for evaluating the performance of the system. During development, a separate development dataset was used, while the test dataset used to get the experimental results in Section \ref{sec:exp} was only used after the system was implemented.

\subsection{Data and sources}
The data is split into two main categories: books and comics.
The books used are mainly books targeted at younger audiences, particularly ``light novels'' which is a genre of Japanese literature mainly targeted at young adults. The reason for this is that books written for adults often contain little to no \textit{furigana}. The comics were chosen with no particular emphasis on any one type as most comics use \textit{furigana}. 

The included pages are the first pages from each of the works, that is, the first pages of the main content (leaving out the title page, table of contents, etc). Aside from this selection of the starting point, the pages are selected as an uninterrupted sequence, even if that means including pages with no \textit{furigana}, full-page illustrations, etc.
\par
The full list of works used can be seen in Appendix \ref{app:data}.

\subsection{Annotation}\label{sec:annotation}
The annotations are made using axis-aligned bounding boxes. This is a natural fit for the data since it is taken from scanned pages of printed media, and thus contains little to no warped text.

The dataset is annotated on character level. Most western text detection datasets are annotated on a word-level basis such as \cite{karatzas2015icdar,veit2016coco,ch2017total}, while Yuan et al. annotate their Chinese scene text dataset on a character level \cite{yuan2019large}. \textit{Furigana} is not regular text, so defining annotations based on words or other grammatical structures is not appropriate. Therefore, the least biased and most flexible style of annotation is to annotate on the character level. This also means that the definition of how the annotations should be done is unambiguous, and should thus provide consistent annotations.
\par
The annotations are created by NKB. Due to the simple definition of the annotations, the annotation work could be done by others without knowledge of Japanese text. As a sanity check, NKB asked four people with no knowledge of Japanese and who are not involved in this project, to create annotations for a single page. They all completed this task with no errors in the annotation.    

\begin{figure*}
\centering
\includegraphics[width=0.8\textwidth]{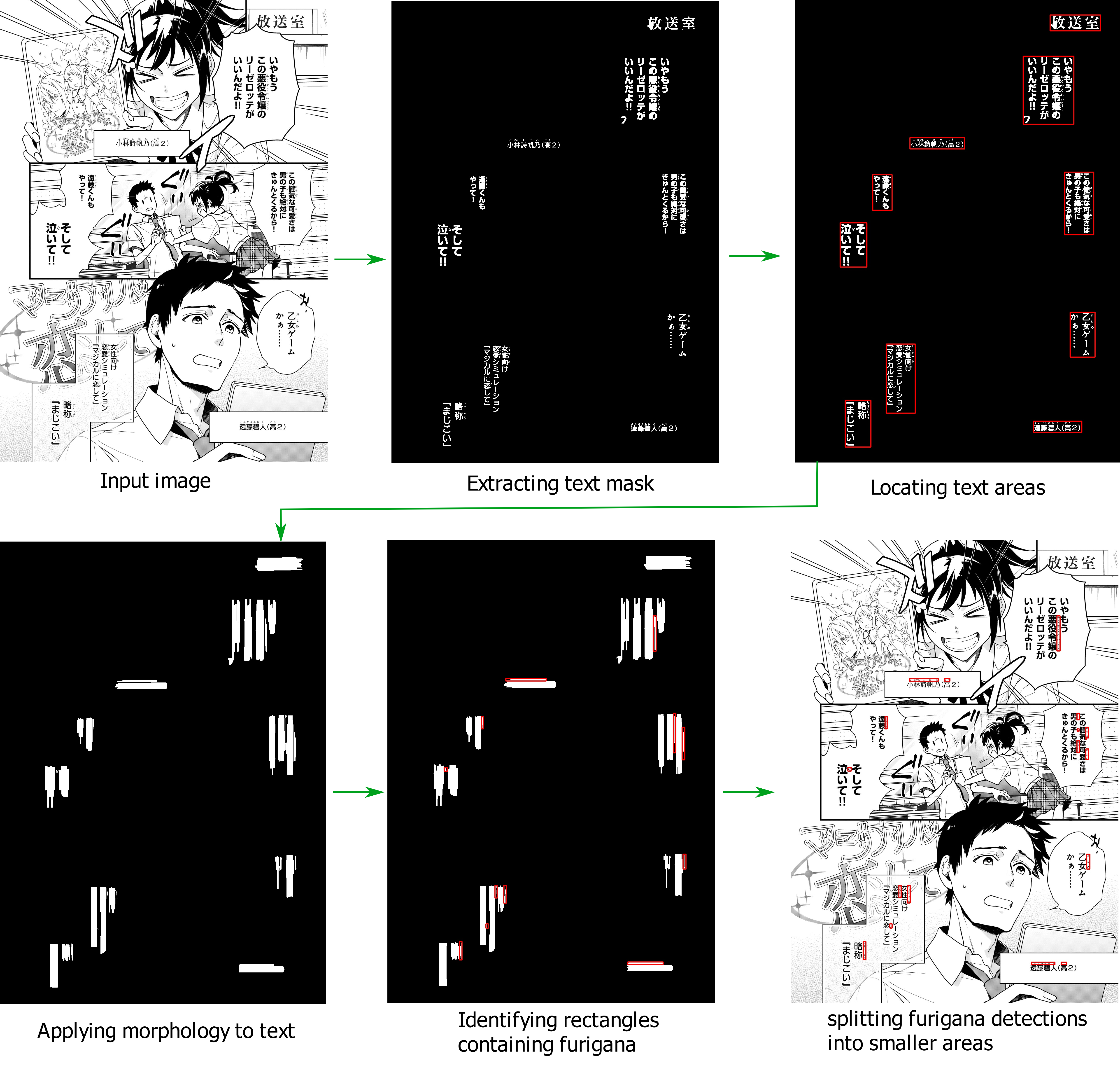}
\caption{Visualization of the detection process. Image source: ``Tsundere Akuyaku Reijō Rīzerotte to Jikkyō no Endō-kun to Kaisetsu no Kobayashi-san'' by Suzu Enoshima\cite{tsundere}.}
\label{process}
\end{figure*}

\section{Method}\label{sec:method}

The proposed method for detecting \textit{furigana} includes the following steps, also visualized in Fig. \ref{process}:
\begin{enumerate}
  \item Extracting text from the image, based on one of the text detection models from Manga Image Translator
  \item Locating text regions, based on morphological operations to merge close-by characters together, and inferring text direction
  \item Separating text lines via morphological operations
  \item Estimating font size of each text line to identify which text is \textit{furigana} 
  \item Splitting \textit{furigana} detections into smaller areas
  \item Optional: verifying \textit{furigana} detections using OCR
\end{enumerate}

The method is composed of these parts because no method exists for \textit{furigana} detection. In our preliminary experiments, scene text detection did not work for detecting text in Japanese media due to a lack of Japanese models, the current top performer in scene text tasks TextFuseNet \cite{ye2020textfusenet} did not work on data from this project. Using OCR did not work well either, we describe this in more detail in Section \ref{sec:experiment:otherocr}. On the other hand, the methods for extracting text from manga worked quite well. We saw good initial results using morphology\footnote{Mathematical morphology involves defining a kernel (or structuring element) consisting of geometrical shape of pixels, e.g. a 2x2 square. This kernel "slides" over a binary image, and depending on how the kernel intersects with pixels in the image, new pixels are created or existing ones are removed. For a good introduction, see \url{https://towardsdatascience.com/understanding-morphological-image-processing-and-its-operations-7bcf1ed11756}} and connected component analysis to detect \textit{furigana} in books. The text extraction methods allowed this approach to also work for comics. We describe more specific implementation details below.

\subsection{Extracting text from the image}\label{sec:method:extracting}
For books, pages consist of mainly text, and the text can be extracted by simply thresholding the image. However, this is not the case for media like comics that contain visual elements in addition to text. To find \textit{furigana} we first need to extract the text while not getting any of the background noise. In this project, we use one of the text detection models from Manga Image Translator by zyddnys and other contributors \footnote{\url{https://github.com/zyddnys/manga-image-translator}}. The section below describes the reasoning for this choice.

\subsubsection{Choice of text extractor}
Text detection is a well-researched problem, Ye et al. \cite{ye2014text} gives an overview of the topic, detailing the individual sub-problems of a text detection system and describing the different methods used to solve these problems, finally giving an overview of some of the proposed systems found in the literature.
\par
There are several large datasets such as \cite{veit2016coco}, \cite{ch2017total} and \cite{karatzas2015icdar} that contain images with annotated text and many text detection models have been proposed based on these datasets. There are a few problems with applying these models directly to the proposed system. Most importantly, these datasets do not contain Japanese text. Also, these datasets contain scene text, that is, text captured "in the wild" and not from written media like books and comics, this means that the regularity that these provide is not exploited. For these reasons, a more specific approach is needed. 
\par
A system capable of detecting text in comics should also be able to solve the relatively easier problem of detecting text in books. Arimaki et al. \cite{aramaki2016text} propose a method for detecting text in manga (Japanese comic books) that considers geometrical features of connected components within the image to generate text region proposals. These regions are then classified using a deep learning model based on the ImageNet\cite{imagenet_cvpr09} classifier by Krizhevsky et al.\cite{krizhevsky2012imagenet} in order to filter away false positives. Their method generates square text regions. Chu et al. \cite{chu2018text} propose a similar method of region proposal and classification. Their method uses a modified Faster R-CNN \cite{ren2015faster} to generate region proposals and classify said regions. Tolle et al. \cite{tolle2011method} propose a method whereby speech balloons in manga are detected using connected component analysis. This means that this method cannot detect text that is not inside a speech balloon. Piriyothinkul et al. \cite{piriyothinkul2019detecting} propose a method where letter candidates are generated using Stroke Width Transform. These candidates are then classified using an SVM trained on the HOG features of the letter proposals. The system developed by Mantra \cite{hinami2020towards} also does text detection in manga but the paper does not go into much detail on how the text extraction is performed.
\par
The results for these methods are of mixed quality, with \cite{piriyothinkul2019detecting} being the top performer with an F1-score of 0.5, although a direct comparison is not possible since different datasets are used for evaluation across these papers. None of the authors provide their source code, so I was unable to test the performance of these papers on the data used in this project.
\par
Del Gobbo \cite{del2020unconstrained,gobbo2020unconstrained} proposes a method, Manga Text Segmentation (MTS), which uses a deep network model with a U-net \cite{ronneberger2015u} architecture to perform end-to-end text detection. Besides good performance, the system outputs a pixel-level mask of the text. This is advantageous, especially in situations where text is placed on top of a background since this background noise will be filtered away. I tested the system developed by Del Gobbo on data from this project. Del Gobbo's system generally performed quite well on many pages. However, a common issue was an abundance of noise/false positives in the output. That is, there are a lot of shapes in the output that is not text. These performance issues might partially stem from the fact that relatively little training data was used, 450 images in tota, compared to, for example, over 60000 images in the COCO-Text dataset \cite{veit2016coco}. 
\par
Besides methods proposed in the literature, there also exist a few other projects that aim to solve this problem. Two are of particular interest. First is Comic Text Detector (CTD) by dmMaze\footnote{\url{https://github.com/dmMaze/comic-text-detector}}. This project, like MTS \cite{del2020unconstrained}, uses an end-to-end deep learning approach. This project however is trained on 13 thousand images of which 1/3 are from the manga109 dataset \cite{aizawa2020building} also used by MTS, 1/3 are from western comics and 1/3 are synthetic data. The other project is Manga Image Translator (MIT). This project provides both its own text detection and one based on CTD. Their original text detector does not work for this project, as the mask is too fuzzy and dilated for the \textit{furigana} to be separated from the main text. The implementation based on CTD on the other hand performs very well, even outperforming the original CTD due to having less noise in the mask. The reason for this difference is unclear since they supposedly use the same model and post-processing.
\\
In conclusion, the best text extractor currently available according to the requirements of this project is MIT (using their implementation based on CTD), and thus this is used for text detection in the proposed method. It should be noted however that each system has their pros and cons. Particularly, there is a trade off between the amount of false positives and false negatives, some systems mark too many shapes as text, and some mark too little. MIT erroneously filters away some of the \textit{furigana}, whereas CTD and MTS generally preserve more of it. But the higher amount of noise in the latter two means that MIT outperforms them. The performance of the different methods are compared in Section \ref{sec:experiment:main}.\\

\subsection{Finding text areas and inferring text direction}\label{sec:method:orientation}
A single page may contain multiple text areas with different orientations and of differing font sizes. In a comic, a text area might equate to one speech bubble and on a book page, there might be a text area for the main text and one for the header.
\par
First, we run a morphological closing operation to join characters that are close to each other. These initial areas are then split into horizontal and vertical groups based on the orientation of the text. This is done by considering the height/width ratio of the text area. If the area is taller than it is wide it is considered vertical, if not it is considered horizontal. This simple yet effective approach has some limitations, which we discuss in Section \ref{sec:experiment:additional}.
\par
Next, areas that are close to each other are merged. This is done for horizontal and vertical groups individually to avoid merging horizontal and vertical text into one area. Doing this improves the performance of the system slightly, as can be seen in Section \ref{sec:experiment:additional}.
\par
The result is a list of rectangles that define text areas as well as information about whether a given text area is horizontal or vertical.

\subsection{Separating text lines}\label{sec:method:morph}
\begin{figure}[!t]
\centering
\includegraphics[width=2.5in]{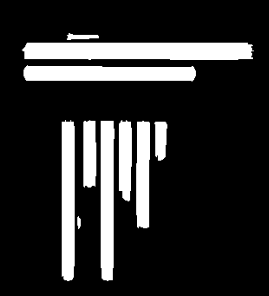}
\caption{Morphology applied to horizontal and vertical text. Closing operation with a long thin kernel. The lines of text are clearly visible. The small blobs next to two of the lines are \textit{furigana}.}
\label{morph}
\end{figure}

In this step, the characters in each line of text are merged to make the entire line appear as a long rectangle. To achieve this, a very long and thin kernel, e.g. a 1/40 ratio, is used while doing a closing operation. The kernel is horizontal for horizontal text and vertical for vertical text. The result is that the text is merged in the direction of the text but not to the sides where the \textit{furigana} is located. An example of this can be seen in Fig. \ref{morph}. The bounding boxes of the connected components are returned as input to the next step.
\par
Before the operation described above is performed, erosion is performed with a small kernel. This is to avoid having the \textit{furigana} touch the main text and accidentally being detected as part of the main text. This is an important step, as not being able to separate \textit{furigana} from the main text is a big source of errors. The difference in performance can be seen in Section \ref{sec:experiment:additional}.

\subsection{Estimating font size and inferring which text is \textit{furigana}}

\begin{algorithm}
\caption{Algorithm for finding font size}\label{alg:font}
\begin{algorithmic}
\State \textbf{Input}: a list of rectangles $R$, a bin size $b$
\State $x \gets$ \textbf{min}(thickness of rectangle in R)
\State $y \gets$ \textbf{max}(thickness of rectangle in R)
\If{$x = y$}\\
\;\;\;\;\Return $x$
\EndIf
\State $font\textunderscore size \gets 0$
\State $max\textunderscore area \gets 0$
\For{$i=x$ to $y$}
    \State $area \gets$ \textbf{sum}(area of rectangles in R with thickness between i and i+b)
    \If{$area \geq max\textunderscore area$}
        \State $max\textunderscore area \gets area$
        \State $font\textunderscore size \gets$ \textbf{mean}(thickness of rectangles in bin)
    \EndIf
\EndFor\\
\Return $font\textunderscore size$
\end{algorithmic}
\end{algorithm}

We now have a list of rectangles that may be either \textit{furigana} or regular text. We can differentiate \textit{furigana} because it is smaller then the regular text. However, in order to generalize this for a diverse set of images, we need an approach that does not assume any absolute sizes. The way this is done is to estimate the font size of the main text.

The process is shown in Algorithm \ref{alg:font}. We assume that the main text has the largest total area of all the shapes. We first get the thickness of all rectangles: the width, if the text is vertical and the height, if the text is horizontal. Then, we find the minimum and maximum thickness. We then divide this interval into bins and distribute the rectangles into them according to their thickness. The bin with the largest total area is assumed to be the bin containing the main text. We then calculate the average thickness of the rectangles within this bin, this is the font size. 

Once we have the font size, inferring which text is \textit{furigana} is quite simple, we define \textit{furigana} as all text rectangles that are less than half the size of the main text, plus a small buffer and a minimum value to avoid marking noise, and return the rectangles that fit this requirement.

\subsection{Splitting detections into smaller areas}\label{sec:method:splitting}
Before returning the final detections, they are split into smaller clusters of characters to avoid marking empty space. Failing to do this often results in evaluation case 9 from Fig.\ref{evaluation}, meaning that the detection is marked as a false positive. Section \ref{sec:experiment:additional} shows the decrease in performance when not splitting the clusters. The splitting is done by running a morphological closing operation on the text within the detection to merge characters that are close to each other but not separated clusters of characters.

\subsection{(Optional) Verifying \textit{furigana} detections using OCR}
We can use OCR to verify the detected \textit{furigana} locations. A weakness of the method described above is that it does not have any concept of language, it is based purely on visual processing. This means that the system does not make any guarantees that the areas selected even contain text. Noise, figures, parts of illustrations, etc. can be selected as \textit{furigana} if it has the right shape and location.
\par
Tesseract is used for OCR operations. While Tesseract has its flaws when it comes to Japanese text, it was by far the best OCR solution out of the ones tested. Paddlepaddle failed to detect any text at all in most cases and OCRopus did not have a compatible Japanese model.
\par
At the end of the previous step, a list of rectangles possibly containing \textit{furigana} is provided. These are used to crop the original image to get a new image containing only the area of interest. Before passing this image to the OCR engine, two prepossessing steps are applied. The image is resized to double size and a white margin is added to the image. This helps improve the performance of Tesseract \footnote{These pre-processing steps are taken from Tesseract's documentation \url{https://tesseract-ocr.github.io/tessdoc/ImproveQuality.html}}. 
\par
The following configurations are applied to Tesseract. Tesseract has two models for Japanese text, one for horizontal text and one for vertical, the correct one is selected based on the orientation of the text. Tesseract allows the user to specify the Page Segmentation Mode (PSM). PSM allows one to specify what kind of text the input image contains. If the text is horizontal, the PSM is set to ``Single Line'', for vertical text it is set to 
``Single Block Vertical Text'' and for square text, assumed to be one character, it is set to ``Single Character''. Finally, using the tessedit\_char\_whitelist variable, the set of accepted characters is limited to  \textit{hiragana} and  \textit{katakana} characters. If no restriction on the set of valid characters is done, there will very often be some character that happens to have high confidence. Also, invalid characters would not be disregarded. There is only one case where this approach fails and that is in detecting \textit{furigana} that is written in \textit{kanji}. However this is very rare, and as described in Section \ref{sec:furigana} it is debatable whether this kind of notation can even be considered \textit{furigana}. 
\par
In order to determine if the image provided is valid, the confidences of the words returned by Tesseract are inspected. If the mean confidence is above a threshold, the image is considered valid. Additionally, if the confidence of any word is above another, higher, threshold, the image is also accepted. If an image contains no characters, or characters not of a correct type, the confidences will be low or there will simply be no detected words. The use of the term ``word'' is somewhat misleading, I found that in most cases the 
``words'' returned by Tesseract corresponded to single characters or groups of two or three characters. This behaviour is understandable given that there is no logical way to split \textit{furigana} into words.
\par
For filtering away invalid detections, this approach works very well as can be seen in Section \ref{sec:experiment:main} from the increase in precision. The main issue with it is that sometimes, word confidences are very low even for valid text. This means that when using this verification step, a lot of correct detections will be filtered away. 

\newcommand\x{3mm}
\newcommand{\xbold}[1]{\setBold[0.6]#1\unsetBold}
\begin{table*}
\caption{Main results. R: Recall(\%). P: Precision(\%). F1: F1-score(\%). Scores marked in bold are the top scores for R/P/F1 within each category. MIT: Manga Image Translator\protect\footnotemark. CTD: Comic Text Detector\protect\footnotemark. MTS: Manga Text Segmentation \cite{gobbo2020unconstrained}.}
\label{tab:eval}
\begin{tabular}{M{30mm}    m{\x} m{\x} m{\x}      m{\x}       m{\x} m{\x} m{\x}      m{\x}       m{\x} m{\x} m{\x}   m{\x}    m{\x} m{\x} m{\x}       m{\x}          m{\x} m{\x} m{\x} }
    \multirow{2}{6em}{Method} & \multicolumn{3}{x{15mm}}{All} & & \multicolumn{3}{x{15mm}}{Books} & & \multicolumn{3}{x{15mm}}{Books (no textbooks)} & & \multicolumn{3}{x{15mm}}{Books (only textbooks)}&  & \multicolumn{3}{x{15mm}}{Comics}\\
    \cline{2-4} \cline{6-8} \cline{10-12}  \cline{14-16} \cline{18-20} 
    & R & P & F1 &   & R & P & F1 &   & R & P & F1 &   & R & P & F1 &   & R & P & F1 \\
    \hline   
    Final method (MIT)                  & \xbold{75} & 83 & \xbold{76} &      & 88 & 84 & 85 &              & \xbold{91} & 92 & 91 &                & 80 & 60 & 67 &                  & \xbold{59} & 82 & \xbold{67} \\
    \hline
    With OCR validation (MIT)      & 66 & \xbold{91} & 73 &        & 83 & \xbold{92} & \xbold{86} &          & 85 & \xbold{95} & 89 &     & 77 & \xbold{84} & \xbold{79} &          & 45 & \xbold{90} & 57 \\
    \hline
    Alternative text detection: thresholding    & 47 & 47 & 46 &                  & 82 & 79 & 79 &                 & \xbold{91} & 94 & \xbold{92} &             & 56 & 38 & 44 &                       & 0 & 0 & 0 \\
    \hline
    Alternative text detection: CTD          & 73 & 78 & 73 &                 & \xbold{89} & 82 & 83 &          & \xbold{91} & 92 & 91 &            & 80 & 54 & 62 &                      & 55 & 72 & 61 \\
    \hline
    Alternative text detection: MTS          & 74 & 77 & 73 &                 & 88 & 81 & 82 &          & 87 & 84 & 84 &                  & \xbold{91} & 74 & 78 &                  & 57 & 71 & 61 \\
    \hline
\end{tabular}
\label{tab:multicol}
\end{table*}

\section{Evaluation metric}\label{sec:evaluation}

\begin{figure}
\centering
\includegraphics[width=2.5in]{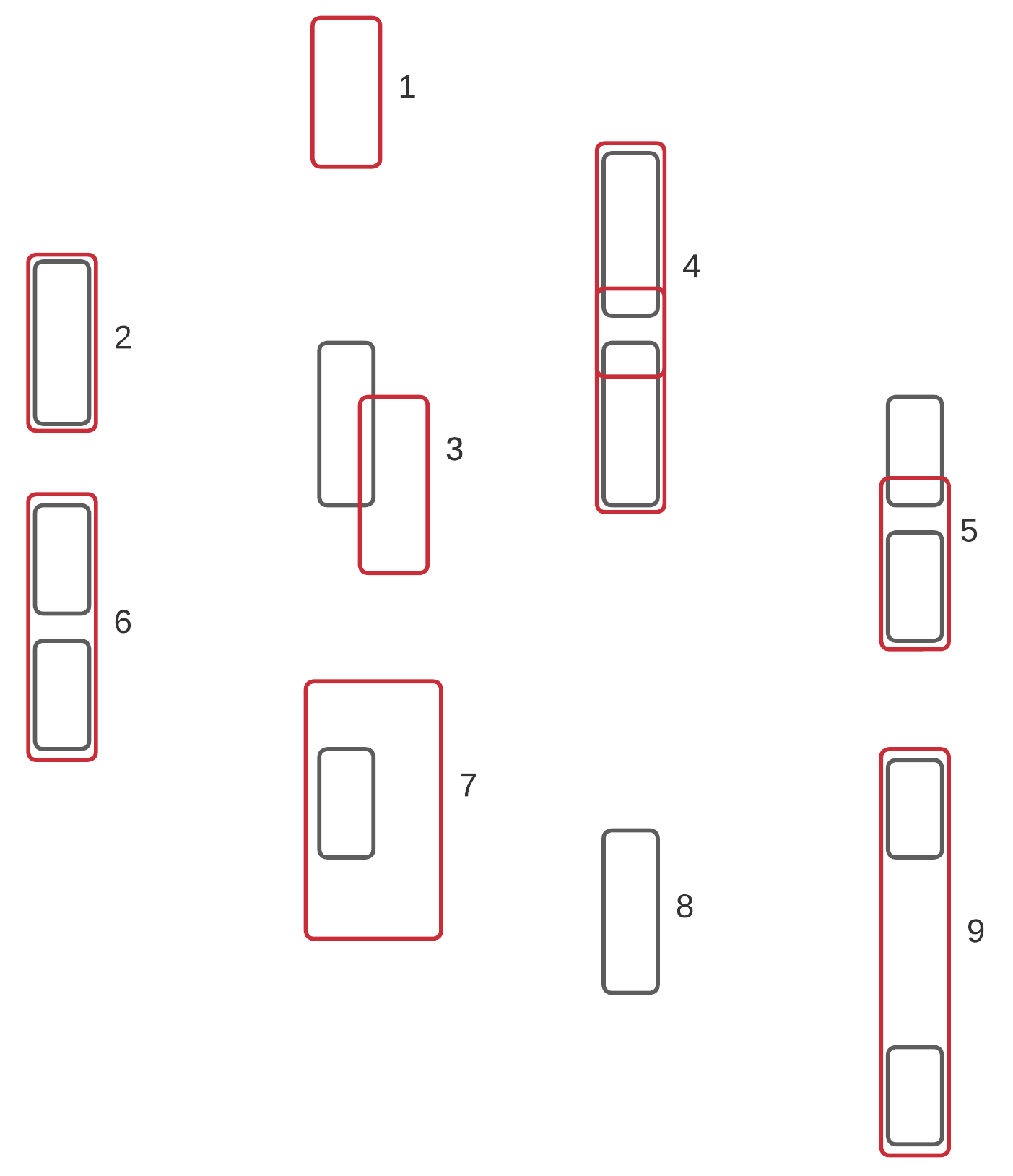}
\caption{Different cases for evaluation. The red boxes should be seen as the predictions made by the system and the black boxes are the ground truth. 1: FP, detection not on ground truth. 2: TP, detection fits ground truth. 3: FP, detection does not sufficiently cover ground truth. 4: TPx2, detections overlap, but sufficiently cover ground truth. 5: TP+FN, detection overlaps one ground truth annotation but does not sufficiently overlap the second one. 6: TPx2, detection overlaps two ground truth annotations. 7: FP, detection area too large. 8: FN, no detection. 9: FP, detection area too large.}
\label{evaluation}
\end{figure}

For evaluating the performance of the system, we use an approach similar to that used in most object detection challenges. Each bounding box detected by the system is compared to the bounding boxes of the annotation, if the detection is similar enough to one of the annotations, the detection is evaluated as a true positive (TP). Otherwise, it is evaluated as a false positive (FP). These counts of TP and FP can then be used to calculate a recall, precision, and F1-score.  
This approach to evaluation is illustrated in Fig. \ref{evaluation}.

\begin{figure}
\centering
\includegraphics[width=2.5in]{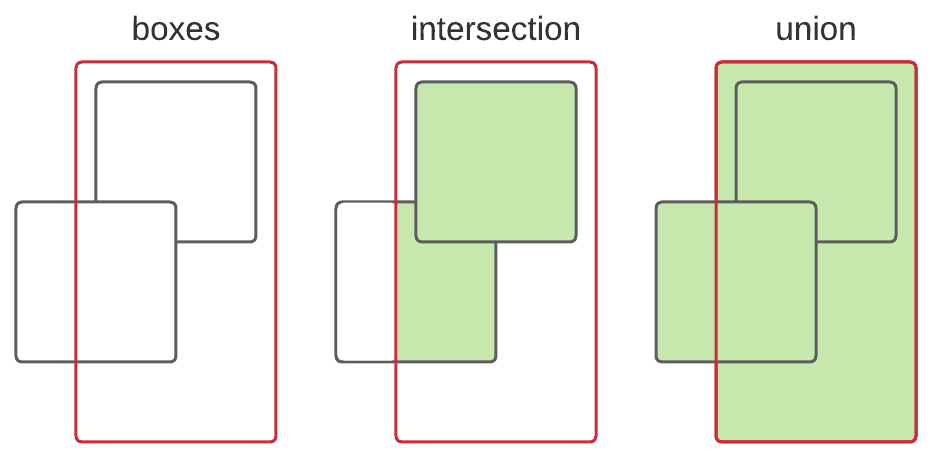}
\caption{Example of intersection and union in n-IOU}
\label{niou}
\end{figure}

This protocol is very similar to the Intersection Over Union (IOU) metric used in most object detection evaluation algorithms such as \cite{padilla2020survey}, as well as the evaluation used by Microsoft COCO \cite{lin2014microsoft}. In the IOU metric, whether a detection sufficiently overlaps a ground truth label is decided by looking at the IOU between the detection and the label. The protocol described in Fig. \ref{evaluation} differentiates itself only in case number 6. Here the detection covers two ground truths. In usual object detection systems, this is an error since the two ground truths should be detected as two distinct objects. However, in the case of detecting \textit{furigana}, there is only one class of object. Additionally, there is no need to separate distinct objects. Since the goal is simply to detect all furigana, it should not matter how/if the characters are grouped together as long as there is not excessive space between grouped characters. 
\par
To facilitate this difference, we introduce the concept of IOU between multiple objects, henceforth referenced to as n-IOU. Usually, IOU is only defined between two objects, a good description of this can be found in  \cite{rezatofighi2019generalized}, however this does not fit the method of evaluation defined above. In Fig. \ref{niou} the n-IOU between the red box and the two black boxes is illustrated. The intersection is the area where the red box intersects with either of the black boxes, the union is the union of all three boxes. The concept can be extrapolated to any amount of black boxes. Mathematically, n-IOU is defined as
\begin{equation}
\frac{R\cap(B_1\cup B_2\cup \dots \cup B_n)}{R\cup B_1\cup B_2 \cup \dots \cup B_n}
\end{equation}

This definition can be extrapolated to work with any amount of R sets as well, but this is not used in this project. 


In case 6 in Fig \ref{evaluation}, the IOU is not large enough between either of the two black boxes and the red box for them to count as a true positive, however the n-IOU between the two black boxes and the red box is large enough. This is scored as two true positives, one for each ground truth. Using n-IOU has the benefit of abstracting away the exact format of the annotations and predictions, meaning that the clustering of characters or lack thereof does not impact the evaluation. 

\subsection{Implementation of the evaluation function}
The evaluation protocol is implemented as follows. For each prediction, the system will check for overlapping labels. If there is one or more overlapping labels, the IOU or n-IOU between the predictions and the label(s) is calculated. If this is above the threshold defined when running the evaluation, the number of true positives is increased by the amount of matching labels. The standard threshold of 0.5 commonly used in object detection is also the default here. This is done for all combinations of the overlapping labels. The combination with the most labels (the best fit), that is within the IOU threshold, is selected. The matching labels are added to a list of used labels and cannot be used again for other predictions. From these counts of true and false positives, the recall, precision, and f1-score are calculated. If true positives + false positives equals 0, precision is undefined and the image is skipped. If true positives + false negatives equals 0, recall is undefined and the image is skipped.

\section{Experiments}\label{sec:exp}

We evaluated the proposed method in several ways:
\begin{itemize}
    \item Performance of the method according to the main evaluation protocol (Section \ref{sec:experiment:main})
    \item Quantitative and qualitative evaluation of the method's errors (Section \ref{sec:experiment:errors})
    \item Assessing the improvement in OCR performance using the proposed method (Section \ref{sec:experiment:ocr} )
    \item Additional experiments with alternative configurations of the method (Sections \ref{sec:experiment:otherocr} and \ref{sec:experiment:additional})
\end{itemize}

\subsection{Main experiment: performance of system according to evaluation protocol}\label{sec:experiment:main}
\addtocounter{footnote}{-2}
\stepcounter{footnote}\footnotetext{\url{https://github.com/zyddnys/manga-image-translator}}
\stepcounter{footnote}\footnotetext{\url{https://github.com/dmMaze/comic-text-detector}}

We present the main results in Table \ref{tab:eval}. Next to splitting the results into columns corresponding to books and comics, we also show two sub-categories of books: regular books and textbooks used to teach Japanese. We can see that both the recall and especially the precision is quite low for textbooks. Whereas Japanese books generally follow a predictable layout, these textbooks often have a complex layout and often include English. Please note that the number of textbook pages is smaller than the other categories. 

The rows show different versions of the method. The first row shows the proposed method which is using MIT as text detection. On the whole dataset, an F1-score of 76 is achieved. MIT is especially good for comics where it beats the other text detection methods. This leads to it beating the other methods in overall F1-score, but it see, it does not hold top scores in all areas.   

The second row shows the results with the optional step of validating detections using OCR enabled. As expected, this increases precision at the cost of recall. The type of data that sees the biggest improvement when using validation is textbooks. This is due to the high level of noise and false detections for this data type. This improvement in score for textbooks means that the F1 score across all books is higher when using validation, although the score for only regular books is slightly lower. 

The third row shows results for the 'thresholding' text detection method. This is simply thresholding the entire image without using any model for extracting text. For regular books, we can see that this method works quite well, reaching an F1-score of 92. Another advantage of the thresholding approach is that it is much faster than the other text detection methods. With the thresholding method, the entire dataset is processed in about 10 seconds on the computer used for testing whereas MIT and CTD take about 150 seconds to process the dataset. However, the thresholding method struggles with textbooks, and is completely useless for comics, getting an F1-score of 0. 

The fourth row shows the results from CTD text detection. CTD's main strength is regular books. But for comics, CTD does not perform as well as MIT due to having more noise in the mask.

The fifth row shows the results from MTS. Overall, MTS also sees weaker results compared to MIT. However, MTS greatly outperforms the other text detection methods in textbooks. There are two main reasons for this. One, MTS generally includes more text in its mask. Whereas the other text detection methods often fail to include some text, especially \textit{furigana}, this is not as common for MTS. Two, MIT and CTD include some noise present in the scanned pages. MTS does not include this noise.

\subsection{Error types}\label{sec:experiment:errors}

\begin{figure*}
\centering
\caption{Examples of error types. The red boxes show \textit{furigana} detections, except for (9) where it shows the detected text area. The images with black background show the text mask.}
\label{fig:errortypes}
\captionsetup[subfigure]{labelformat=empty}
    \begin{subfigure}[t]{0.21\textwidth}
             \fbox{\includegraphics[width=\textwidth]{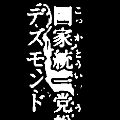}}
             \caption{(1) Error in text mask, there are two lines of text but among the characters is a lot of noise, most noticeably the white blob starting from the middle of the two lines and extending to the right line. Src: \cite{spy}}
    \end{subfigure}
    \hspace{1em}
    \begin{subfigure}[t]{0.21\textwidth}
             \fbox{\includegraphics[width=\textwidth]{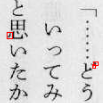}}
             \caption{(2) Disconnected bits of a characters detected as \textit{furigana}. Src:\cite{kido}}
    \end{subfigure}
    \hspace{1em}
    \begin{subfigure}[t]{0.21\textwidth}
             \fbox{\includegraphics[width=\textwidth]{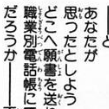}}
             \caption{(3) \textit{Furigana} close to main text, which results in the \textit{furigana} not being separated from the main text and thus not being detected. Src:\cite{peppermint}}
    \end{subfigure}
    \hspace{1em}
    \begin{subfigure}[t]{0.21\textwidth}
             \fbox{\includegraphics[width=\textwidth]{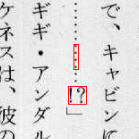}}
             \caption{(4) Special characters detected as \textit{furigana}. Src:\cite{kido}}
    \end{subfigure}
    
    \vspace{0.2cm}
    
    \begin{subfigure}[t]{0.42\textwidth}
             \centering
             \fbox{\includegraphics[width=\textwidth]{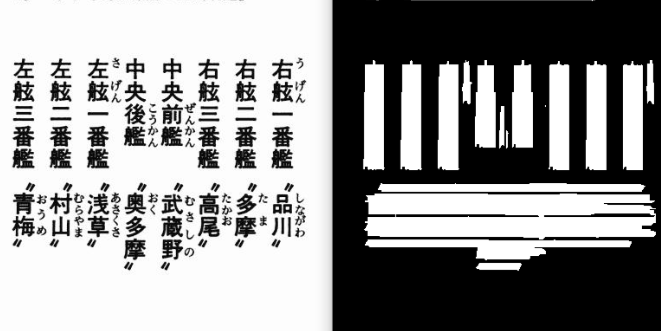}}
             \caption{(5) Wrong orientation detected. The original text is shown on the left and the text mask with morphology applied to it on the right. The text on the top is correctly detected as being vertical, but the text on the bottom is detected as being horizontal which results in the text lines being "stretched" in the wrong direction. Src:\cite{kyo}}
    \end{subfigure}
    \hspace{4em}
    \begin{subfigure}[t]{0.42\textwidth}
             \centering
             \fbox{\includegraphics[width=\textwidth]{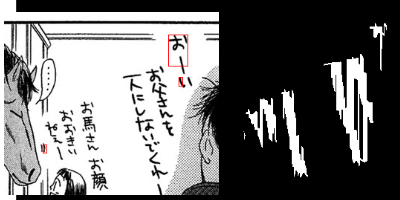}}
             \caption{(6) text with non-standard formatting. On the right is the original image. On the left is the image with morphology applied, here we can see what goes wrong. Some of the lines are merged together into one large blob, smaller blobs next to this will be detected as \textit{furigana}. Src:\cite{derby}}
    \end{subfigure}
    
    \vspace{0.2cm}
    
    \begin{subfigure}[t]{0.2\textwidth}
             \centering
             \fbox{\includegraphics[width=\textwidth]{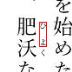}}
             \caption{(7) \textit{Furigana} too small. The first two \textit{furigana} are correctly detected, but the last one is so thin that it disappears when processing the image. Src:\cite{nito}}
    \end{subfigure}
    \hspace{3em}
    \begin{subfigure}[t]{0.20\textwidth}
             \centering
             \fbox{\includegraphics[width=\textwidth]{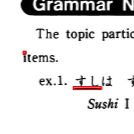}}
             \caption{(8) English text causing problems. The dot above the i in "items" is detected as \textit{furigana}. Src: \cite{jpfor}}
    \end{subfigure}
    \hspace{3em}
    \begin{subfigure}[t]{0.40\textwidth}
             \centering
             \fbox{\includegraphics[width=\textwidth]{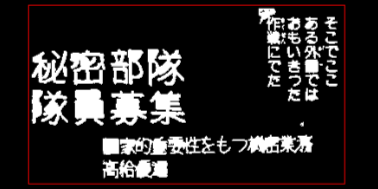}}
             \caption{(9) Wrong text area estimation. Here we see three text areas, however since they are very close the system detects them as one area. The red box shows the detected area. Src:\cite{peppermint}}
    \end{subfigure}
\end{figure*}

Table \ref{tab:errors} shows the categories of errors within the detections of the test data. The errors are enumerated both by the number of pages that a given error type appears on and by how many errors in total are caused by this error type. For example, error type 2 appears on 17 pages but only causes 26 errors in total, while error type 5 causes over 112 errors on only 6 pages. E.g. when the wrong text orientation is detected, all the \textit{furigana} within that block of text is not detected and creates many false negatives. Figure \ref{fig:errortypes} shows examples of these errors.

\begin{table}
\caption{Error types}
\label{tab:errors}
\begin{tabular}{m{0.12\linewidth}  M{0.5\linewidth}  m{0.09\linewidth}  m{0.09\linewidth}}
Error nr. & Error description & Pages & Total \\
\hline
1 & Errors in text mask & 33 & 260   \\
\hline
2 & Disconnected bits of a character detected as \textit{furigana}  & 17 & 26 \\
\hline
3 & \textit{Furigana} is so close to main text that it is not correctly separated from it & 12 & 173 \\
\hline
4 & Special characters (exclamation mark, underline etc.) is detected as \textit{furigana} & 8 & 44  \\
\hline
5 & Wrong text orientation detected & 6 & 112  \\
\hline
6 & Text with non-standard formatting (slanted, hand-written etc.)  & 6 & 49 \\
\hline
7 & \textit{Furigana} is too small  & 3 & 3 \\
\hline 
8 & English text & 2 & 2 \\
\hline
9 & Incorrect text area estimation & 1 & 16 \\
\end{tabular}
\end{table}

\subsection{Improvement in the performance of OCR}\label{sec:experiment:ocr}

One of the goals of detecting \textit{furigana} was to improve the performance of OCR of Japanese text. To test this with the proposed method, we used the manga109 dataset, which includes both locations and annotations for text blocks. 

The experiment was performed as follows. For each page, the \textit{furigana} is detected and removed by replacing the \textit{furigana} area with white pixels. Then, for each text block, OCR (Tesseract) is performed on the text block in the original image and the version with the \textit{furigana} removed. For each of these, the Levenshtein edit-distance\cite{levenshtein1966binary}, and based on this the Character Error Rate ($distance/n$), is calculated. We can then compare the two to see if there has been an improvement in the CER. 

\begin{table}
\caption{Performance of OCR with and without \textit{furigana} removal. Performance is given in CER (lower is better).}
\label{tab:ocrimp}
\begin{tabular}{ M{0.3\linewidth}  m{0.09\linewidth}  m{0.11\linewidth}  m{0.14\linewidth} m{0.11\linewidth} }
Data/method & text blocks & CER with \textit{furigana} & CER \textit{furigana} removed & improve-ment \\
\hline
first 15 books of manga109 & 20893 & 0.496 & 0.472 & +5\%   \\
\hline
first 15 books of manga109, books with frequent \textit{furigana} usage & 15345 & 0.510 & 0.476 & +7\%   \\
\hline
first 15 books of manga109, books with infrequent \textit{furigana} usage & 5548 & 0.459 & 0.461 & -0.3\%   \\
\hline
Akkera Kanjinchou, \textit{furigana} removed with proposed method & 1340 & 0.532 & 0.473 & +11\%   \\
\hline
Akkera Kanjinchou, \textit{furigana} removed manually & 1340 & 0.532 & 0.450 & +15\%   \\
\end{tabular}
\end{table}

We show the results in Table \ref{tab:ocrimp}. Removing \textit{furigana} using the proposed method improves the performance of OCR slightly. The amount of improvement varies based on the book. The largest improvement was observed in Belmondo with an improvement of 12\% and the smallest improvement was seen in Burari Tessen Torimonocho, where the CER decreased by 1.6\%. 

One thing to note about these results is that not all the manga in manga109 include \textit{furigana}. In cases where \textit{furigana} is not used, trying to remove \textit{furigana} can only negatively affect the performance, since everything that is removed is a false positive. The second and third row show this difference in performance. 

It is also of interest to compare removing \textit{furigana} using the proposed method to removing \textit{furigana} manually as a "gold standard" \textit{furigana} removal. To test this, we manually removed the \textit{furigana} from the Akkera Kanjinchou manga. Removing \textit{furigana} using the proposed method sees an improvement of 11\%, while removing \textit{furigana} manually improves the performance by 15\%. In other words, the proposed method reaches 72\% of the maximum possible improvement in performance that can be achieved by removing \textit{furigana}.

\subsection{Alternative configurations - detecting \textit{furigana} using OCR}\label{sec:experiment:otherocr}

\begin{figure}[!t]
\centering
\includegraphics[width=0.4\textwidth]{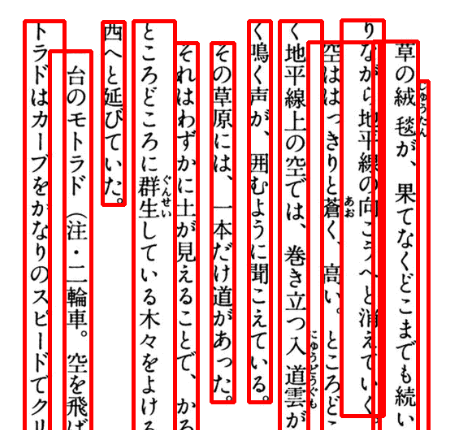}
\caption{Text lines in a book as detected by tesseract. Src:\cite{kino}}
\label{textlines}
\end{figure}

An interesting alternative to the proposed method would be to use OCR to detect what text is \textit{furigana}. The first step of most OCR algorithms is to detect the location of text. In theory, we could cut out many of the error-prone steps in the proposed method by using these detections as the starting point for the detection of \textit{furigana}. If we had the bounding boxes for the main text and the \textit{furigana}, it would be simple to compare the sizes and determine which is \textit{furigana}.

In practice, this approach has some major problems. The assumption that we can get accurate text locations with main text and \textit{furigana} does not hold. To test this approach, we used Tesseract to get text locations. Tesseract often fails to detect main text and \textit{furigana} as separate lines, which creates inaccurate text lines as well as making it impossible to use this information to separate \textit{furigana} and main text.

These issues with text lines can be seen in Figure \ref{textlines}. The \textit{furigana} is correctly separated from the main text in the first line, but not so in any of the following lines. This was the case for all the images tested. The same issues are present at the word level bounding boxes, where words often contain both \textit{furigana} and main text (Tesseract is not made to give accurate character level annotations, so these are not accurate regardless). 

This result also gives some insight into why \textit{furigana} worsens the output of OCR systems. The \textit{furigana} makes the system detect text lines incorrectly, which in the worst case means that the main text will not be detected correctly since the \textit{furigana} next to the characters might cause misclassifications.

\subsection{Other alternative configurations}\label{sec:experiment:additional}

Table \ref{tab:additional} shows the results of changing various details of the method. The first row is the main result from Section \ref{sec:experiment:main}, while rows 2 to 4 show the results of switching off parts of the method described in Sections \ref{sec:method:orientation}, and \ref{sec:method:splitting}.

The last row presents another additional experiment with using OCR to detect the direction of the text. Inferring the direction of the text is a non-trivial problem. The approach in Section \ref{sec:method:orientation} fails if, for example, we have a long row of short vertical lines in an area. Then the area is longer than it is tall, but the lines are still vertical. A better approach is to look at the height/width ratio of the individual lines. This can be done by using an OCR engine (here Tesseract) to find the lines of text within the text area. The number of vertical and horizontal lines is counted, if there are more vertical lines, the area is considered vertical and vice versa. However, this method is much more computationally expensive. And as we can see, it ends up with worse performance than the simpler baseline method. 

\begin{table}
\caption{Experiments using different configurations of the method.}
\label{tab:additional}
\begin{tabular}{ M{0.3\linewidth}  m{0.15\linewidth}  m{0.15\linewidth} m{0.15\linewidth} }
Experiment & Recall(\%) & Precision(\%) & F1(\%) \\
\hline
Baseline & 75 & 83 & 76 \\
\hline
Not merging text areas & 74 & 87 & 77\\
\hline
Not using erosion during morphology step & 63 & 86 & 68\\
\hline
No splitting of \textit{furigana} areas & 46 & 80 & 54 \\
\hline
Using OCR to detect text orientation & 75 & 82 & 76\\
\end{tabular}
\end{table}

\section{Discussion}\label{sec:discussion}
\subsection{Discussion of results}
In Section \ref{sec:experiment:main}, we saw that the system performs better on books than on comics. The dataset is a mix of newer and older comics, and most of the errors seem to be related to older comics. In older comics, the \textit{furigana} is written very close to the main text. Due to this and the often poor resolution of older comics and it becomes hard to separate the main text from the \textit{furigana}. This is the main reason for the low recall seen on comics as it means that much of the \textit{furigana} is not detected. For newer comics, the scores are more comparable to books.
\par
Section \ref{sec:experiment:ocr} showed that it is possible to improve the performance of OCR by removing \textit{furigana}. One thing that stands out is that the error rate is quite high, even with the \textit{furigana} removed manually. This indicates that \textit{furigana} is not the main problem with Tesseract's OCR. Other OCR services show better results than this. For example, using Google's OCR solution seems to make much fewer errors when transcribing Japanese text, despite the fact that it is based on Tesseract. This shows that there should be ample room for improvement in the general performance of the OCR. If the error rate was lower overall, it is possible that the relative improvement seen from removing \textit{furigana} would be higher than the improvements seen in Section \ref{sec:experiment:ocr}. A big contributor to the high error rate is that when the detected text for a single text area is longer than the expected text, the error rate for this text area will be very high. As an example, in one text area the expected text is "!!?", but Tesseract outputs "\includegraphics[height=3mm]{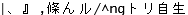}", in this case, the character error rate is 5. These kinds of detections considerably increase the error rate. 
\par
Section \ref{sec:experiment:additional} showed the results of the OCR-based method of inferring text orientation. The method generally works well, but text lines are not always detected correctly. Adding some heuristics for when the OCR-based method should be used, we were able to achieve a performance that was slightly better than the main results. However, given that this method significantly slows down the system, the lack of strong improvement, and the general inconsistency of the method, we decided to not use this method of inferring text orientation.

\subsection{Future work}
The performance of the method depends in large part on the performance of the auxiliary problem of extracting Japanese text from book and comic pages into a text mask. The best text detection system for the proposed method is Manga Image Translator. This problem could use more attention, as the current methods are still far from perfect.
\par
One issue with the proposed method is that it does not start by considering the text itself, a human reads a text one character at a time and can use its position, size and meaning relative to other characters to determine its role in the text, while the proposed method sees only connected components of pixels. A method that is integrated into an OCR system would be able to use this textual information, but would require an OCR engine capable of fine-grained processing on the character level.
\par
Another possibility would be to modify the text line detection of an OCR engine such that it would recognize \textit{furigana} as distinct lines. This would make it possible to implement a simple \textit{furigana} detection mechanism based on the location and size of the detected lines.\\

\section{Conclusion}
Detection of \textit{furigana} in Japanese written media is a problem that has not seen much research. This project formally defines the problem, discusses the related challenges and proposes a method for detecting the locations of \textit{furigana} in Japanese books and comics based on morphology and connected component analysis. For evaluating this method, we created a new dataset of images with annotations of \textit{furigana}, and proposed an adjusted detection metric. Additionally, we show that using this method can improve the performance of OCR in texts where \textit{furigana} is used frequently. The method performs well on pages of regular layout but struggles with unusual or complex text layout. While comics and textbooks might require more fundamental changes to the method, it's likely that with some fine-tuning this method can achieve near-perfect results on regular Japanese books.

\bibliographystyle{IEEEtran}
\bibliography{refs}

\onecolumn
\appendices  
\section{Dataset}\label{app:data}

Table \ref{tab:data} lists the books and comics included in the test dataset. We do not have the rights to share all the publishers images, however, the images and annotations (including the development set with 57 images) can be provided on request by contacting \href{mailto:nikolajnkb@gmail.com}{\texttt{nikolajnkb@gmail.com}}.

\begin{table*}[!htbp]
\caption{Dataset}
\label{tab:data}
\centering
\begin{tabular}{P{0.2\linewidth}  P{0.2\linewidth}  P{0.1\linewidth}  P{0.1\linewidth}  P{0.1\linewidth}  P{0.08\linewidth} }
Title          & Romanized title   & Author   & Publisher   & Nr. of pages     & Pages used    \\
\hline & \\[-1.5ex]
\multicolumn{5}{c}{Books} \\
\hline & \\[-1.5ex]
ようこそ実力至上主義の教室   & Yōkoso Jitsuryoku Shijō Shugi no Kyōshitsu e   & Shōgo Kinugasa 
& Media Factory    & 5    & book 3, page 10-14  \\
\hline & \\[-1.5ex]
境界線上のホライゾン    & Kyōkaisen-jō no Horaizon   & Minoru Kawakami   
& ASCII Media Works & 5    & book 1, page 22-26   \\
\hline & \\[-1.5ex]
新機動戦記ガンダムW Frozen Teardrop & Shin Kidō Senki Gandamu Wingu Furozen Tiadoroppu & Katsuyuki Sumisawa 
&    Kadokawa Shoten   &    5         &         book 1, page 8-12 \\
\hline & \\[-1.5ex]
魔法科高校の劣等生 & Mahōka Kōkō no Rettōsei & Tsutomu Satō & Shōsetsuka ni Narō & 5 & book 1, page 12-16\\
\hline & \\[-1.5ex]
ニートだけどハロワにいったら異世界につれてかれた & Nītodakedo Harowa ni Ittara Isekai ni Tsuretekareta & Katsura Kasuga & Media Factory & 5 & book 1, page 10-14\\
\hline & \\[-1.5ex]
ガーリー・エアフォース & Gārī Ea Fōsu & Kōji Natsumi & ASCII Media Works & 5 & book 1, page 12-17\\
\hline & \\[-1.5ex]
上級へのとびら & Jōkyu e no Tobira & Mayumi Oka et al. & Kurosio Publishers & 5 & page 166-170\\
\hline & \\[-1.5ex]
Japanese for everyone & Japanese for Everyone & Susumu Nagara & Gakken & 5 & page 86-90\\
\hline & \\[-1.5ex]
\multicolumn{5}{c}{Comics} \\
\hline & \\[-1.5ex]
ボン・クラージュ！乙女 & Bon Kurāju! Otome & Akisato Wakuni & Shogakukan & 5 & book 1, page 7-11\\
\hline & \\[-1.5ex]
ダービークイーン & Dābīkuīn & Ashihara Hinako & Shogakukan & 5 & book 1, page 5-10 \\
\hline & \\[-1.5ex]
カミヨメ & Kami Yome & Tomiyaki Kagisora  & Square Enix & 5 & book 1, page 5-9\\
\hline & \\[-1.5ex]
ミステリと言う勿れ & Misuteri to Iu Nakare & 	Yumi Tamura & Shogakukan & 5 & book 1, page 6-10\\
\hline & \\[-1.5ex]
Spy × Family & Spy × Family & Tatsuya Endo & Shueisha & 5 & book 7, page 5-10\\
\hline & \\[-1.5ex]
屋根裏部屋の公爵夫人 & Yaneura Beya no Koshaku Fujin & Fumino Mori & Kadokawa Game Linkage & 5 & book 1, page 5-9\\
\hline & \\[-1.5ex]
ペパミント・スパイ & Pepaminto supai & Noriko Sasaki & Hakusensha & 5 & book 1, page 6-10\\
\hline
\end{tabular}
\end{table*}

\end{CJK}
\end{document}